\documentclass{article}

\PassOptionsToPackage{numbers, compress}{natbib}
 \usepackage[preprint]{neurips_2025}

\usepackage[utf8]{inputenc} 
\usepackage[T1]{fontenc}    
\usepackage{hyperref}       
\usepackage{url}            
\usepackage{booktabs}       
\usepackage{amsfonts}       
\usepackage{nicefrac}       
\usepackage{microtype}      
\usepackage{xcolor}         
\usepackage{multirow}
\usepackage{amsmath} 
\usepackage{subcaption}
\usepackage{graphicx}

\title{Probing the Limits of Stylistic Alignment in Vision-Language Models}

\author{%
  Asma Farajidizaji \\
  Imperial College London \\
  \texttt{asma.farajidizaji24@imperial.ac.uk} \\
  \And
  Akash Gupta \\
  Apta AI \\
  \texttt{akash@apta.chat} \\
  \AND
  Vatsal Raina \\
  Apta AI \\
  \texttt{vatsal@aptaai.com} \\
}

\begin{document}

\maketitle

\begin{abstract}
Vision-language models are increasingly used to generate image captions in specific styles, such as humor or romantic. However, these transformer-based models often struggle with this subjective task in a zero-shot setting. While preference data can be used to align them toward a desired style, such data is expensive to acquire, limiting the ability to explore the models' full capabilities. This work addresses this by studying the data efficiency of aligning small vision-language models to humor and romantic styles. This approach helps to define the performance limits of these models and determine how little preference data is needed to achieve stylistic saturation, benchmarking their capabilities and limitations.
\end{abstract}

\section{Introduction}

Stylistic image captioning~\cite{osman2023survey} tests whether vision–language models (VLMs)~\cite{zhang2024vision} can control \emph{how} a description is written, not only \emph{what} is depicted~\cite{chen2018factual,mohamad2019towards,yang2020better,zhao2020memcap}. Styles such as humor or romantic are subjective and under-specified, and current transformer VLMs often miss the intended tone in zero-shot settings. Stylistic captioning therefore provides a practical setting to examine alignment methods.
Preference-based alignment optimizes models to match human comparisons or ratings of candidate captions~\cite{ye2025painting,xu2024altogether}. Methods used in language modeling, and adopted here, include supervised fine-tuning on preferred samples (SFT)~\cite{harada2025massive,hua2024intuitive}, policy-gradient approaches such as PPO~\cite{ziegler2019fine}, and direct preference objectives such as DPO~\cite{rafailov2023direct}, with recent variants including MPO~\cite{wang2024enhancing}, ORPO~\cite{hong2024orpo}, and SimPO~\citep{meng2024simpo}. Much of this work has targeted general behavioral goals (e.g., helpful, honest, harmless) in text-only models~\cite{ziegler2019fine}; this study applies the same family of objectives to subjective style in multimodal captioning. In parallel, lighter-weight control techniques such as prompt engineering~\cite{sahoo2024systematic}, soft prompting (e.g., prefix tuning)~\cite{li2021prefix} and linear “steering vectors’’~\cite{cao2024personalized}, offer cheaper knobs for style, though their reliability in multimodal settings is less clear.
As high-quality multimodal preference data are costly, data efficiency is a central concern~\cite{zhang2024vickreyfeedback}. Concretely, the study benchmarks zero-shot and SFT baselines against SimPO on a small VLM across New Yorker cartoons~\cite{hessel2023androids} and Flickr humor/romantic splits~\cite{Gan_2017_CVPR} under controlled preference-data budgets. Evaluation reports win rates against a reference policy, along with validated style-classifier accuracy, to trace data-efficiency curves and identify saturation points. This work asks: \textit{How data-efficient is stylistic alignment for VLMs, and where do improvements saturate?} The contribution is a reproducible protocol and comparative results under explicit data budgets that clarify when gains arise from additional preference data versus model capacity.

\section{Aligning VLMs for Stylistic Generation}

This work assumes preference triplets pairing each image with a stylized caption (positive) and a factual caption (negative), and conditions a vision–language model on the image and an optional style instruction. As a no-training baseline, zero-shot prompting inserts a brief style instruction at inference. Supervised fine-tuning (SFT) imitates only the positive captions to teach the target style, while direct preference objectives exploit the positive–negative contrast without reward models or on-policy rollouts. In particular, SimPO increases the length-normalized log-probability of positives relative to negatives. 
A held-out test set provides image–caption triplets (image, factual, stylized). For evaluation, \textbf{WR-LogP} counts a win when the model’s length-normalized log-probability for generating the target stylized caption exceeds that for the factual caption. \textbf{Style-Acc} is the accuracy of a binary style classifier applied to the model’s stylized generations i.e whether the binary style classifier classifies the generated caption as the desired style.

\section{Results}

\begin{figure}[htbp]
  \centering
  \begin{subfigure}{0.32\textwidth}
    \centering
    \includegraphics[width=\linewidth]{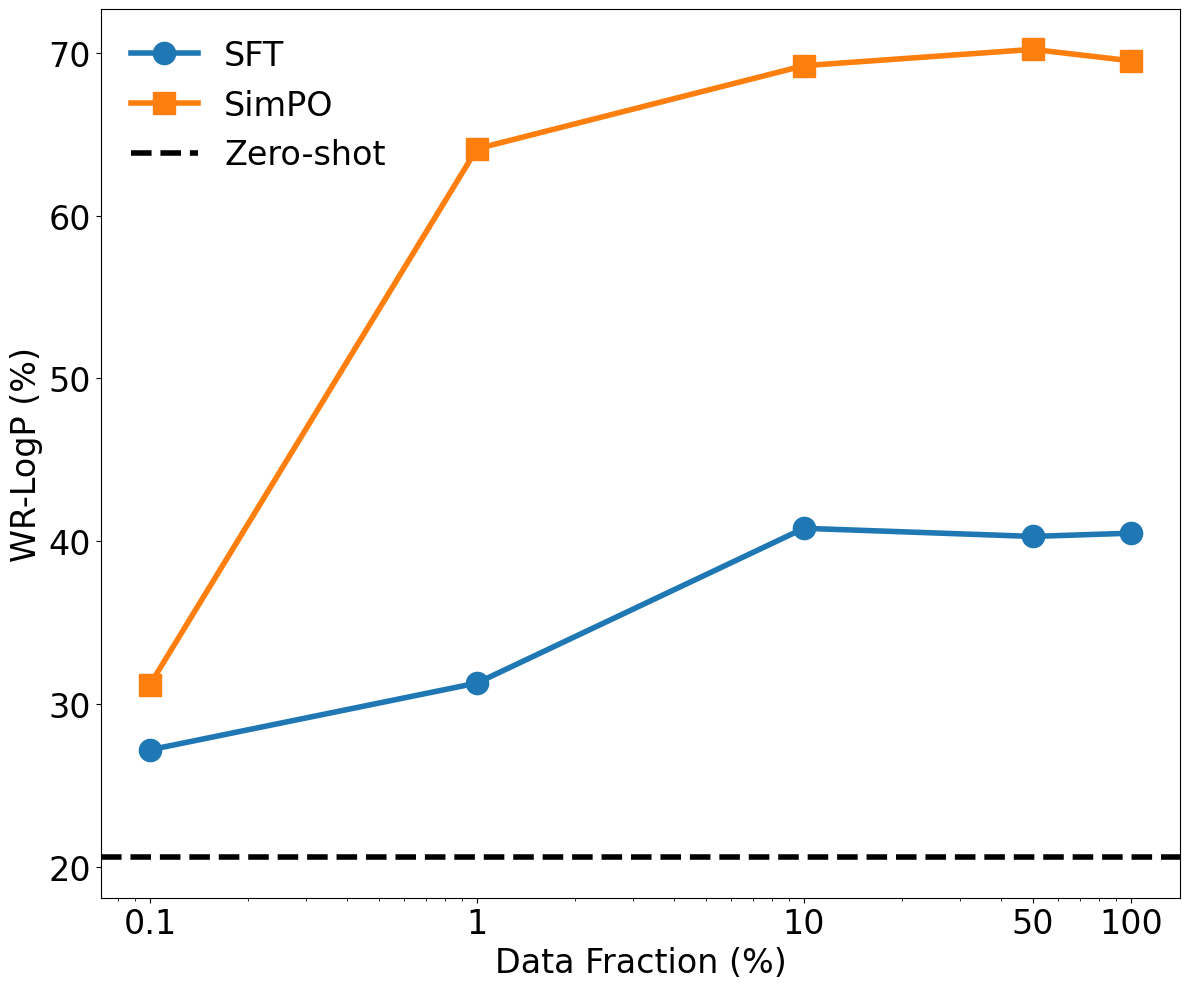}
    \caption{New Yorker}
    \label{fig:newyorker}
  \end{subfigure}
  \hfill
  \begin{subfigure}{0.32\textwidth}
    \centering
    \includegraphics[width=\linewidth]{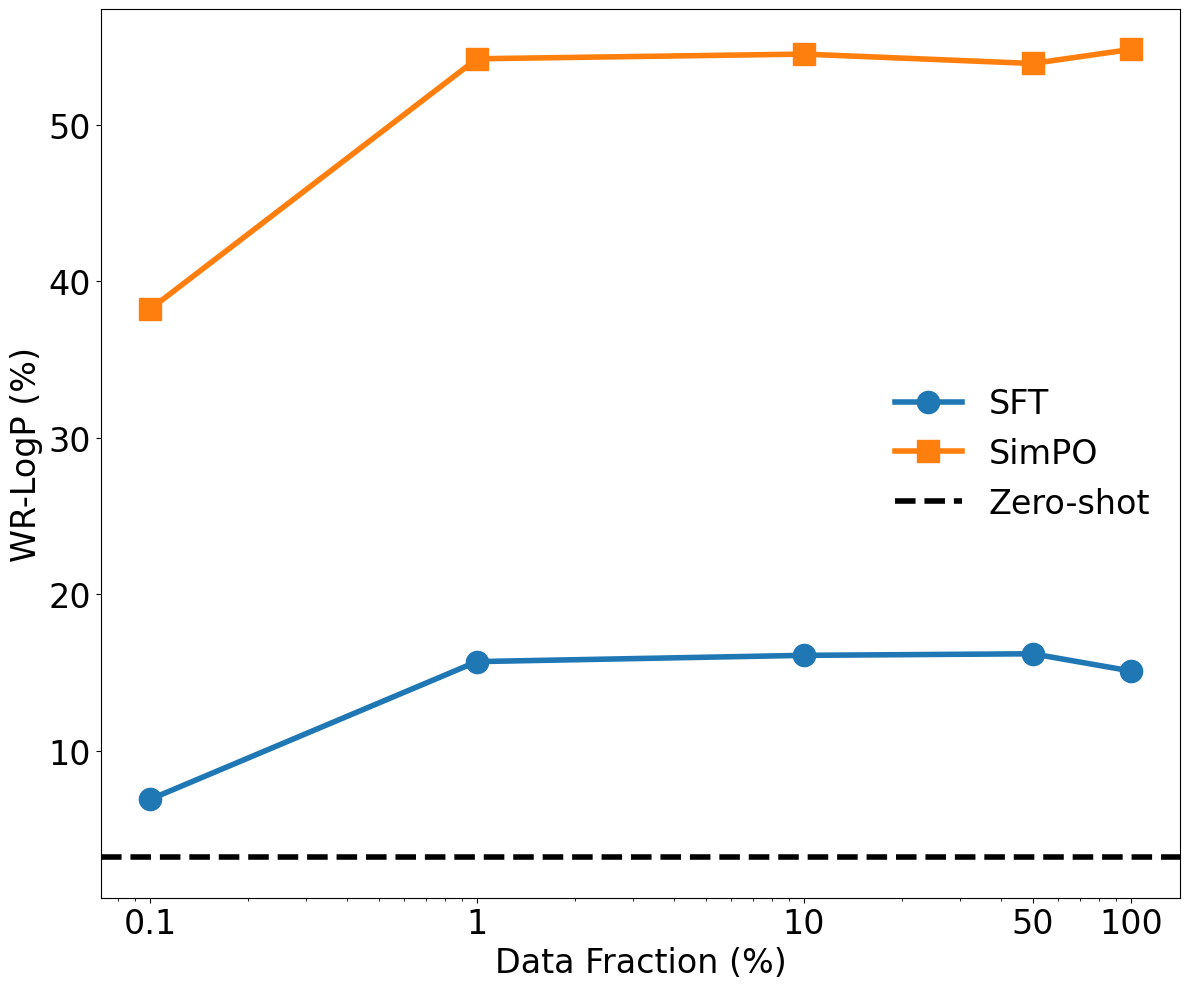}
    \caption{Flickr - humor}
    \label{fig:flickr_humor}
  \end{subfigure}
  \hfill
  \begin{subfigure}{0.32\textwidth}
    \centering
    \includegraphics[width=\linewidth]{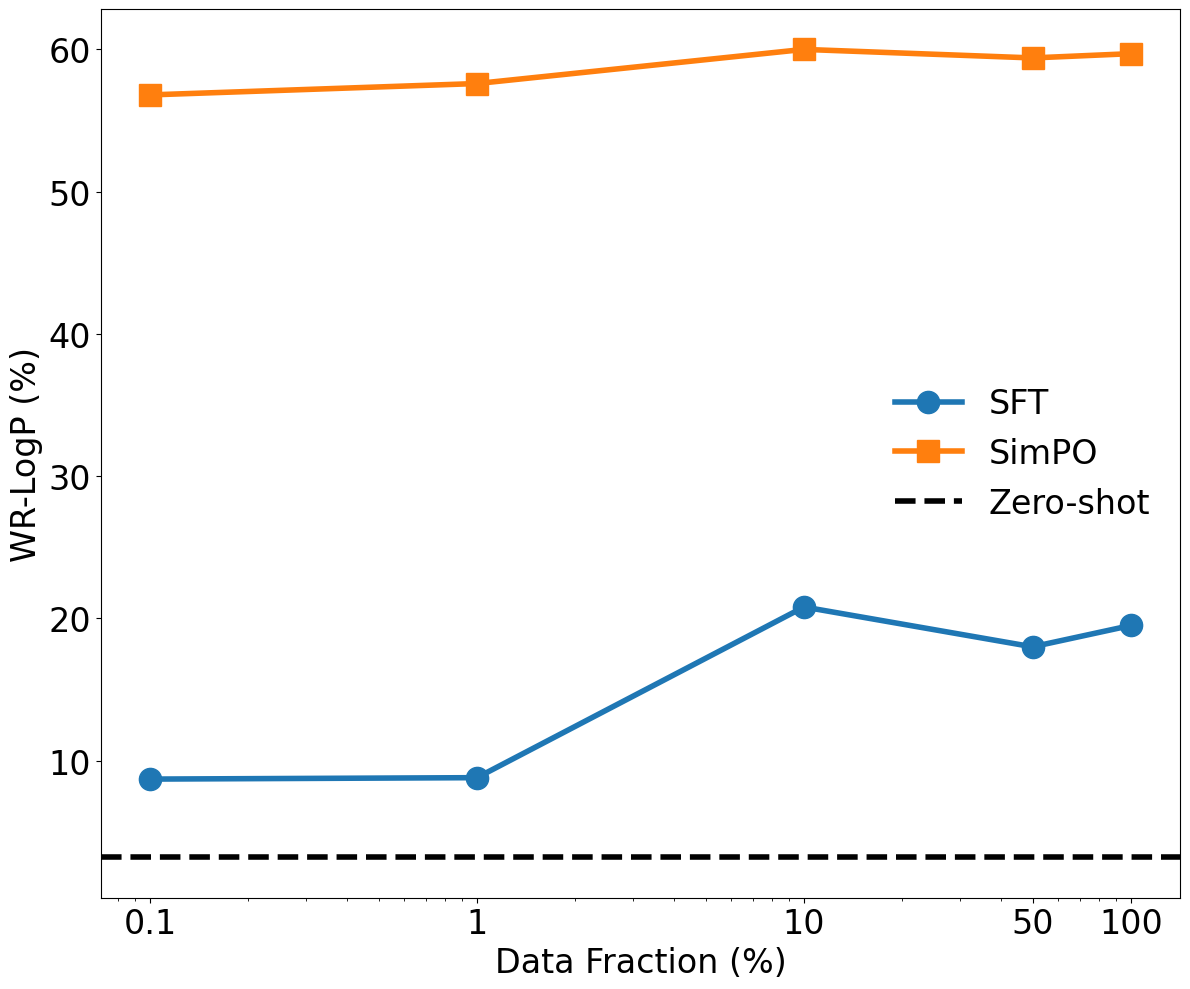}
    \caption{Flickr - romantic}
    \label{fig:flickr_romantic}
  \end{subfigure}
  \caption{Data efficiency curves for stylistic alignment.}
  \label{fig:data_efficiency}
\end{figure}

This work evaluates stylistic alignment on two datasets. The New Yorker Caption dataset~\cite{hessel2023androids} provides 2,340 training, 130 validation, and 131 test examples, each pairing an image with a factual and a humor caption. The FlickrStyle10k dataset~\cite{Gan_2017_CVPR} includes 10k images, of which 6k are used for training (5.4k train, 600 validation) and 1k are used for testing. Each image has seven captions: five factual, one humor, and one romantic, supporting separate alignment tasks for humor and romance. The experiments use Qwen-2.5-VL-3B-Instruct~\cite{bai2025qwen2}, a recent small-scale VLM considered state-of-the-art for its size.
Across both datasets, zero-shot prompting performs worst, SFT yields consistent improvements, and SimPO achieves the strongest alignment (Table~\ref{tab:style_alignment}). The model shows higher gains on the New Yorker dataset, where humor is more formulaic, compared to Flickr humor and romantic captions, which are subtler and harder to capture. Data-efficiency curves (Figure~\ref{fig:data_efficiency}) reveal that improvements saturate quickly, with as little as 10\% of preference data sufficient to reach peak performance. Beyond this point, additional data provides little benefit, suggesting model capacity rather than data volume is the limiting factor for stylistic alignment.

\begin{table}[htbp]
  \centering
  \small
  \caption{Style alignment for image captioning with 100\% preference data for training evaluated using the win rate on the preferred style captions and classification accuracy on the generated captions.}
  \label{tab:style_alignment}
  \begin{tabular}{l *{3}{c c}}
    \toprule
    \multirow{2}{*}{Method} & \multicolumn{2}{c}{\textbf{New Yorker}} & \multicolumn{2}{c}{\textbf{Flickr - humor}} & \multicolumn{2}{c}{\textbf{Flickr - romantic}} \\
    \cmidrule(lr){2-3} \cmidrule(lr){4-5} \cmidrule(lr){6-7}
    & WR-LogP & Style-Acc & WR-LogP & Style-Acc & WR-LogP & Style-Acc \\
    \midrule
    Zero-Shot & 20.6 & 57.3 & 3.2 & 11.0 & 5.4 & 0.6 \\
    SFT & 40.5 & 91.6 & 15.1  & 46.6 & 19.5 & 43.3 \\
    SimPO & 69.5 & 87.8 & 38.8 & 97.5 & 59.7 & 41.9 \\
    \bottomrule
  \end{tabular}
\end{table}

\section{Conclusion}

This work shows that small vision–language models can be aligned to produce captions in subjective styles such as humor and romance, but the task remains challenging. Zero-shot prompting is insufficient, SFT improves performance, and preference-based methods like SimPO provide the strongest gains. However, alignment saturates quickly, indicating that model capacity, rather than data availability, constrains stylistic generation.

\bibliographystyle{plainnat} 
\bibliography{references}


\appendix

\section{Technical Appendices and Supplementary Material}

\subsection{Style Classifier}

The style classifier accuracy is one of the core evaluation metrics reported in the main paper. For each dataset and target style, a binary style classifier was trained on the training split and evaluated on the held-out test split to validate its suitability as an evaluation tool for assessing the generated captions from the aligned models. 

The classifier architecture is lightweight: CLIP embeddings~\cite{radford2021learning} are extracted independently for the input image and the candidate caption. These two embeddings are concatenated, then passed through a two-layer feedforward head for the New Yorker dataset and a four-layer feedforward head for the Flickr dataset with GELU activations~\cite{hendrycks2016gaussian} followed by a final sigmoid output layer. At inference time, the classifier outputs the probability of the caption belonging to the target style, and a threshold of 0.5 is used to make a binary decision.

Table~\ref{tab:style_classifier} reports the precision, recall, F1 score, and accuracy for each dataset on the test split. The strong performance across datasets demonstrates that the classifier generalizes well, providing a reliable and consistent metric for benchmarking the stylistic ability of caption generation models.

\begin{table}[htbp]
  \centering
  \small
  \caption{Binary style classifier accuracy on test splits for each dataset.}
  \label{tab:style_classifier}
  \begin{tabular}{l|ccc|c}
    \toprule
    Dataset & Precision & Recall & F1 & Accuracy  \\
    \midrule
New Yorker & 85.7 & 96.2 & 90.6 & 90.1  \\
Flickr - humor & 85.4 & 83.0 & 84.2 & 84.4 \\
Flickr - romantic & 91.0 & 83.8 & 87.3 & 87.8 \\
    \bottomrule
  \end{tabular}
\end{table}

\subsection{Hyperparameter Details}

All models were trained using A100 or L4 GPUs and took fewer than 2 hours to train each. In general, the validation loss was used for early-stopping and for hyperparameter tuning.

Table~\ref{tab:hyperparams} shows the hyperparameters used for supervised finetuning (SFT). To maintain consistency between the SFT runs, we used the same seed to split train and validation sets, as well as the seed used when initializing weights and training the model. However, we used up to 3 seeds to sample the truncated train set.

\begin{table}[htbp]
  \centering
  \small
  \caption{Hyperparameter details for vision-language model training.}
  \label{tab:hyperparams}
  \begin{tabular}{ll|ccc}
    \toprule
    Training strategy &  & New Yorker & Flickr - humor & Flickr - romantic  \\
    \midrule
    \multirow{4}{*}{SFT} 
      & learning rate & 1.0e-5 & 1.6e-5 & 0.8e-5  \\
      & batch size    & 16     & 16     & 16      \\
      & scheduler     &   linear decay    & linear decay & linear decay \\
      & max steps     & 270    & 600    & 600     \\
    \midrule
    \multirow{4}{*}{SimPO} 
      & learning rate & 2e-5 & 2e-5 & 2e-5  \\
      & batch size    &   32   &   32   &   32    \\
      & scheduler     & cosine & cosine  & cosine \\
      & max steps     &   66  &  170   &   170   \\
    \bottomrule
  \end{tabular}
\end{table}

At inference time, all vision-language models had a generation temperature set to 0.7, max number of tokens to 128 and a beam size of 1 (greedy search).

For the training of the style classifiers, a maximum of 20 epochs was considered, with a binary cross-entropy loss, a learning rate of 2e-4 and a batch size of 32.



\subsection{Limitations}

This work has several potential limitations. First, the experiments are restricted to a small vision–language model; it remains an open question how stronger models behave under the same data budgets and whether their performance saturates differently. Second, only two alignment methods are explored beyond zero-shot prompting: supervised fine-tuning and SimPO. Other recent preference-based objectives, such as ORPO and related variants, may offer complementary insights. Finally, the training setup assumes access to both images and stylized captions. While this provides a richer signal, much prior work considers alignment with only stylized text available~\cite{guo2019mscap}, making the task more challenging but also less resource intensive.

\subsection{Licenses}

The New Yorker Caption dataset is made available under the Creative Commons Attribution 4.0 (CC‑BY‑4.0) license. The FlickrStyle10K dataset is made available under the Creative Commons Attribution-NonCommercial 4.0 International (CC BY-NC 4.0) license.




\end{document}